\documentclass{article}

\usepackage[final]{neurips_2019_ml4ps}
 
\usepackage{graphicx}
\usepackage[colorlinks,urlcolor=blue,citecolor=blue,linkcolor=blue]{hyper ref}
\usepackage{subcaption}
\usepackage{enumitem}
\usepackage{graphicx}
\usepackage{commath}
\usepackage{amsmath}
\usepackage{amssymb}
\usepackage{mathrsfs}
\usepackage{mathtools}
\usepackage{booktabs}
\usepackage{cleveref}
\usepackage{comment}
\usepackage{color, colortbl}
\definecolor{Gray}{gray}{0.9}

\DeclareMathOperator*{\argmax}{arg\,max}
\newcommand\eureqa{\texttt{eureqa}}

\let\vec\mathbf
\begin{document}

\title{Learning Symbolic Physics with Graph Networks}

\author{%
  Miles D.~Cranmer \\
  Princeton University \\
  Princeton, NJ, USA \\
  \texttt{mcranmer@princeton.edu} \\
  \And
  Rui Xu \\
  Princeton University \\
  Princeton, NJ, USA \\
  \texttt{ruix@princeton.edu} \\
  \And
  Peter Battaglia \\
  DeepMind \\
  London, UK \\
  \texttt{peterbattaglia@google.com} \\
  \And
  Shirley Ho \\
  Flatiron Institute \\
  New York City, NY, USA \\
  \texttt{shirleyho@flatironinstitute.org} \\
}

\maketitle

\begin{abstract}
    We introduce an approach for imposing physically motivated inductive biases on graph networks to learn interpretable representations and improved zero-shot generalization. Our experiments show that our graph network models, which implement this inductive bias, can learn message representations equivalent to the true force vector when trained on n-body gravitational and spring-like simulations. We use symbolic regression to fit explicit algebraic equations to our trained model's message function and recover the symbolic form of Newton's law of gravitation without prior knowledge. We also show that our model generalizes better at inference time to systems with more bodies than had been experienced during training. Our approach is extensible, in principle, to any unknown interaction law learned by a graph network, and offers a valuable technique for interpreting and inferring explicit causal theories about the world from implicit knowledge captured by deep learning.
\end{abstract}

\section{Introduction}
Discovering laws through observation of natural phenomenon is the central challenge of the sciences. Modern deep learning also involves discovering knowledge about the world but focuses mostly on \textit{implicit} knowledge representations, rather than \textit{explicit} and interpretable ones. One reason is that the goal of deep learning is 
often optimizing test accuracy and learning efficiency in narrowly specified domains, while science seeks causal explanations and general-purpose knowledge across a wide range of phenomena. Here we explore an approach for imposing physically motivated inductive biases on neural networks, training them to predict the dynamics of physical systems and interpreting their learned representations and computations to discover the symbolic physical laws which govern the systems. Moreover, our results also show that this approach improves the generalization performance of the learned models.

The first ingredient in our approach is the ``graph network'' (GN) \citep{battaglia2018relational}, a type of graph neural network \citep{scarselli2009graph,bronstein2017geometric,gilmer2017neural}, which is effective at learning the dynamics of complex physical systems \citep{battaglia2016interaction,chang2016compositional,sanchez2018graph,mrowca2018flexible,li2018learning,kipf2018neural}. We impose inductive biases on the architecture and train models with supervised learning to predict the dynamics of 2D and 3D n-body gravitational systems and a hanging string. If the trained models can accurately predict the physical dynamics of held-out test data, we can assume they have discovered some level of general-purpose physical knowledge, which is implicitly encoded in their weights. Crucially, we recognize that the forms of the graph network's message and pooling functions have correspondences to the forms of force and superposition in classical mechanics, respectively.
The message pooling is what we call a ``linearized latent space:'' a vector space where latent representations of the interactions between bodies (forces or messages) are linear (summable). By imposing our inductive bias, we encourage the GN's linearized latent space to match the true one.
Some other interesting approaches for learning low-dimensional general
dynamical models include \cite{packard1980geometry,daniels_automated_2015}, and \cite{jaques_2019}.

The second ingredient is using symbolic regression --- we use \eureqa\ from \cite{schmidt2009distilling} --- to fit compact algebraic expressions to a set of inputs and messages produced by our trained model. \eureqa\ works by randomly combining mathematical building blocks such as mathematical operators, analytic functions, constants, and state variables, and iteratively searches the space of mathematical expressions to find the model that best fits a given dataset. The resulting symbolic expressions are interpretable and readily comparable with physical laws.

The contributions of this paper are:
\begin{enumerate}[leftmargin=30pt,labelindent=5pt,itemindent=0pt,noitemsep,nosep]
\item A modified GN with inductive biases that promote learning general-purpose physical laws.
\item Using symbolic regression to extract analytical physical laws from trained neural networks.
\item Improved zero-shot generalization to larger systems than those in training.
\end{enumerate}

\section{Model}

Graph networks are a type of deep neural network which operates on graph-structured data.
The format of the graphs on which GNs operate is defined as 3-tuples\footnote{We adhere closely to the notation used in \cite{battaglia2018relational} to formalize our model.}, $G=(\mathbf{u}, V, E)$, where:
\begin{enumerate}[label={},leftmargin=30pt,labelindent=5pt,itemindent=-15pt,noitemsep,nosep]
    \item $\mathbf{u} \in \mathbb{R}^{L^u}$ is a global attribute vector of length $L^u$,
    \item $V=\{\mathbf{v}_i\}_{i=1:N^v}$ is a set of node attribute vectors, $\mathbf{v}_i \in \mathbb{R}^{L^v}$ of length $L^v$, and
    \item $E=\{(\mathbf{e}_k, r_k, s_k)\}_{k=1{:}N^e}$ is a set of edge attribute vectors, $\mathbf{e}_k \in \mathbb{R}^{L^e}$ of length $L^e$, and indices ${r_k, s_k \in \{1{:}N^v\}}$ of the ``receiver'' and ``sender'' nodes connected by the $k$-th edge.
\end{enumerate}

\begin{figure}[h]
    \centering
    \includegraphics[width=0.75\textwidth]{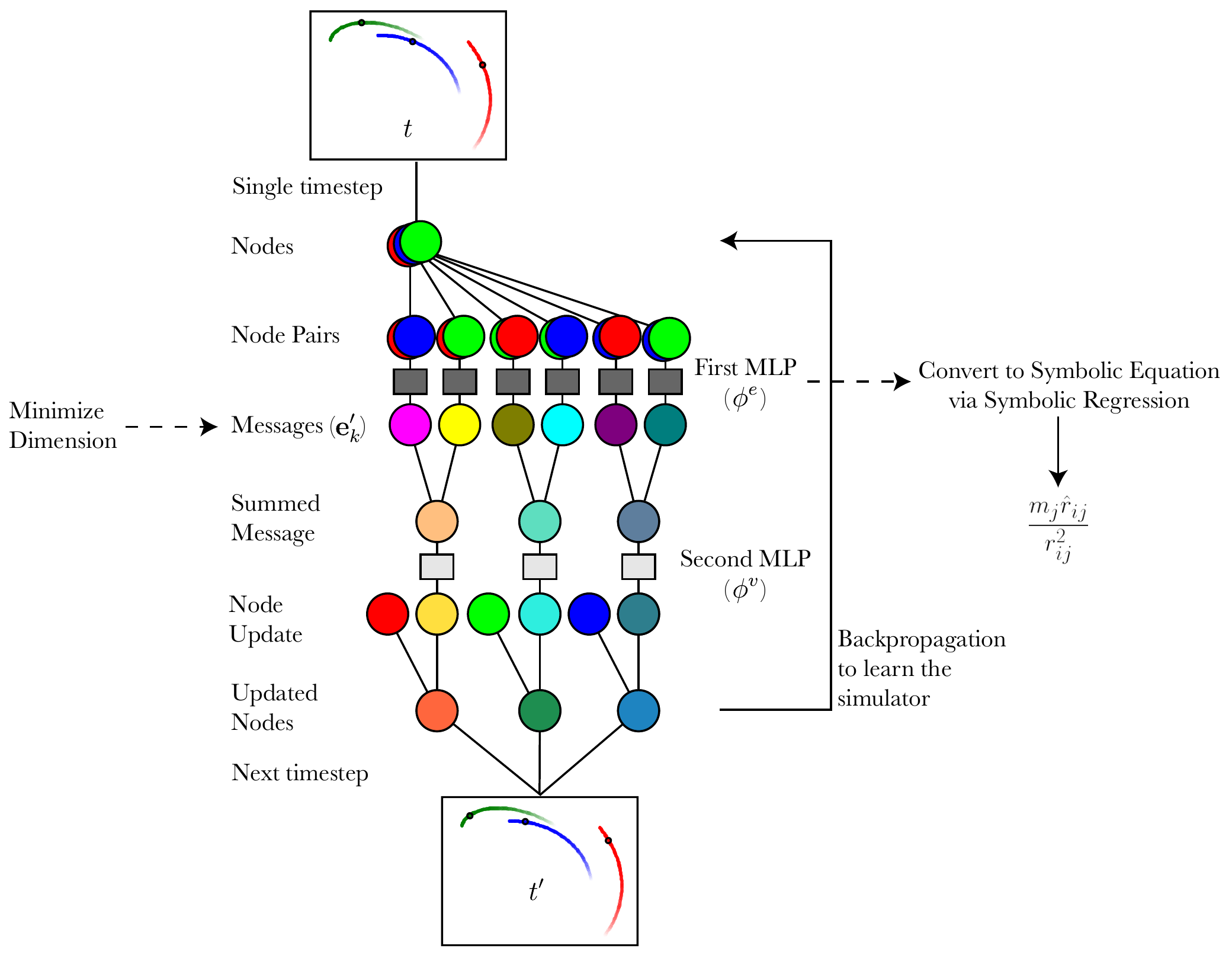}
    \caption{A schematic depicting how we extract physical knowledge from a GN.}
    \label{fig:setup}
\end{figure}

Our GN implementation is depicted in \cref{fig:setup}. Note: it does not include global and edge attributes. This GN processes a graph by first computing pairwise interactions, or ``messages'', $\mathbf{e}'_k$, between nodes connected by edges, with a ``message function'', $\phi^e: \mathbb{R}^{L^v} \times \mathbb{R}^{L^v} \times \mathbb{R}^{L^e} \rightarrow \mathbb{R}^{L^{e'}}$. Next, the set of messages incident on each $i$-th receiver node are pooled into $\mathbf{\bar{e}}'_i = \rho^{e \rightarrow v}(\{\mathbf{e}'_k \}_{r_k = i, k = 1{:}N^e})$, where $\mathbf{\bar{e}}'_i \in \mathbb{R}^{L^{e'}}$, and $\rho^{e \rightarrow v}$ is a permutation-invariant operation which can take variable numbers of input vectors, such as elementwise summation. Finally, the pooled messages are used to compute node updates, $\mathbf{v}'_i$, with a ``node update function'', $\phi^v: \mathbb{R}^{L^v} \times \mathbb{R}^{L^{e'}} \rightarrow \mathbb{R}^{L^{v'}}$. Our specific architectural implementation is very similar to the ``interaction network'' (IN) variant \citep{battaglia2016interaction}.

The forms of $\phi^e$, $\rho^{e \rightarrow v}$, $\phi^v$, and the associated input and output attribute vectors have correspondences to Newton's formulation of classical mechanics, which motivated the original development of INs. The key observation is that $\mathbf{e}'_k$ could learn to correspond to the force vector imposed on the $r_k$-th body due to its interaction with the $s_k$-th body. In our examples, the force vector is equal to the derivative of the Lagrangian: $\frac{\delta \mathcal{L}}{\delta \vec{q}}$, and this could be generally imposed if one knows $\frac{d}{dt}(\frac{\delta \mathcal{L}}{\delta \dot{\vec{q}}})$ and manually integrates the ODE with the output of the graph net.
In a general n-body gravitational system in $n$ dimensions, note that the forces are minimally represented in an $\mathbb{R}^n$ vector space. Thus, if $L^{e'} = n$, we exploit the GN's ``linearized latent space'' for physical interpretability: we encourage $\mathbf{e}'_k$ to be the force. 

We sketch a non-rigorous proof-like demonstration of our hypothesis.  Newtonian mechanics prescribes that force vectors, $\mathbf{f}_k \in \mathcal{F}$, can be summed to produce a net force, $\sum_k \mathbf{f}_k = \mathbf{\bar{f}} \in \mathcal{F}$, which can then be used to update the dynamics of a body. 
Our model uses the $i$-th body's pooled messages, $\mathbf{\bar{e}}'_i$, to update the body's velocity via Euler integration, $\mathbf{v}'_i = \mathbf{v}_i + \phi^v(\mathbf{v}_i, \mathbf{\bar{e}}'_i)$.
 If we assume our GN is trained to predict velocity updates perfectly for any number of bodies, this means $\mathbf{\bar{f}}_i = \sum_{r_k = i} \mathbf{f}_k = \phi^v_i( \sum_{r_k = i}  \mathbf{e}'_k) = \phi^v_i( \mathbf{\bar{e}}'_i)$, where $\phi^v_i(\cdot) = \phi^v(\mathbf{v}_i, \cdot)$.
 We have the result for a single interaction: $\mathbf{\bar{f}}_i = \mathbf{f}_{k, r_k=i} = \phi^v_i(\mathbf{e}'_{k, r_k=i}) = \phi^v_i(\mathbf{\bar{e}}'_i)$. Thus, we can substitute into the multi-interaction case: $\sum_{r_k=i}\phi^v_i(\mathbf{e}'_k) = \phi^v_i(\mathbf{\bar{e}}'_i) = \phi^v_i(\sum_{r_k=i}\mathbf{e}'_k)$, and so $\phi^v_i$ has to be a linear transformation. Therefore, for cases where $\phi^v_i$ is invertible (mapping between the same dimensional space), $\mathbf{e}'_k = (\phi^v_i)^{-1}(\mathbf{f}_k)$, and so the message vectors are linear transformations of the true forces when $L^{e'}=D$.   We demonstrate this hypothesis on trained GNs in \cref{sec:exp}.

\section{Experiments}
\label{sec:exp}

We set up 100,000 simulations with random masses and initial
conditions for both
a $1/r$ and $1/r^2$ force law in 2D, a $1/r^2$ law in 3D, and a string 
with an $r^2$ force law between nodes in 2D with a global gravity, for
1000 time steps each. These laws are chosen arbitrarily as examples
of different symbolic forms. 
The three n-body problems have six bodies in their
training set, and the string has ten nodes, of which the two end
nodes are fixed.
We train a GN on each of these problems where we choose
$L^{e'}=D$, i.e., the length of the message vectors in the GN
matches the dimensionality of the force vectors: 2 for the 2D
simulations and 3 for the 3D simulations.
Our GN, a pure TensorFlow \citep{tensorflow2015-whitepaper} model, has
both $\phi^e$ and $\phi^v$ as
three-hidden-layer multilayer perceptrons (MLPs) with 128 hidden nodes per layer
with ReLU activations. We optimize the L1 loss
between the predicted velocity update and the true velocity
update of each node.

\begin{figure}[h]
    \centering
    \includegraphics[width=0.8\textwidth]{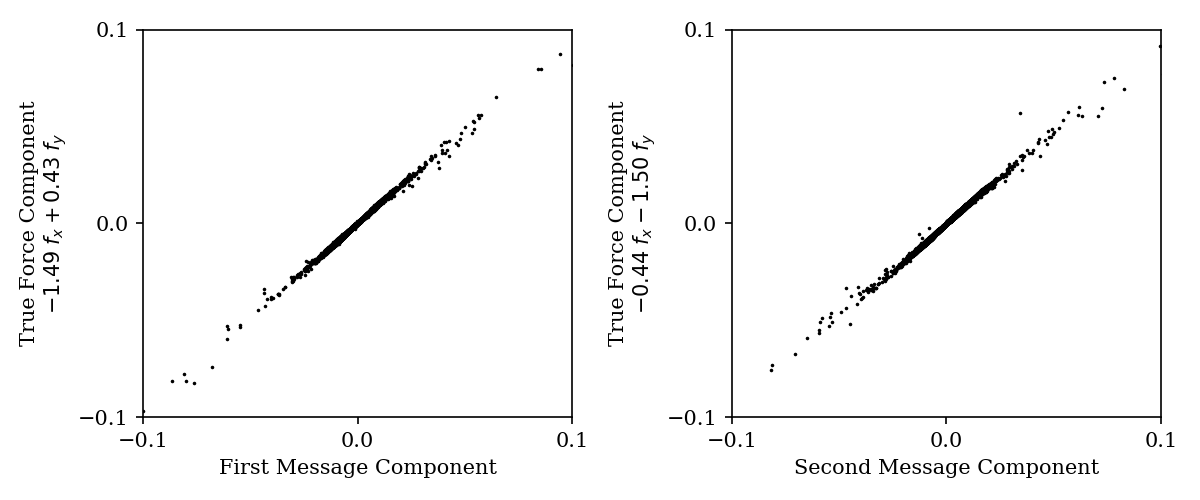}
    \caption{These plots demonstrate
    that the graph network's messages have learned to be linear transformations
    of the two vector components of the true force: $f_x$ and $f_y$, for the $1/r$ law in 2D.}
    \label{fig:square_truth2}
\end{figure}

Once we have a trained model, we
record the messages, $\vec{e}'_k$,
for all bodies
over 1000 new simulations for each environment.
We fit a linear combination of the vector components of
the true force, $\vec{f}_k$, to each component of $\vec{e}'_k$,
as can be seen in \cref{fig:square_truth2} for $1/r$.
The results for each system show that the $\vec{e}'_k$ vectors have learned to be a linear combination of the components when $L^{e'}=D$.
We see similar linear relations for all other simulations.

{}

We are also able to 
find the force law when it is unknown 
by using symbolic regression to fit
an algebraic function that approximates $\phi^e$.
We demonstrate this on the trained GN
for the $1/r$ problem
using \eureqa\ from \cite{schmidt2009distilling}
to fit algebraic equations that fit the message.
We allow it to use algebraic operators
$+, -, \times, /,$ as well as input variables ($\Delta x$
and $\Delta y$ for component separation, $r$ for distance, and $m_2$
for sending body mass)
and real constants. Complexity is scored
by counting the number of occurences of each operator, constant,
and input variable.
This returns a list of the models with the
lowest mean square error at each
complexity. We parametrize
Occam's razor to find the ``best'' algebraic model
by first sorting the best models by complexity,
and then taking the model that maximizes
the fractional drop in mean square error (MSE) over the
next simplest model:
$\argmax_c(-\Delta \log(\text{MSE}_c)/\Delta c)$,
where $c$ is the complexity.
The best model found by the symbolic regression
for the first output element
of $\phi^e$ is $(0.46 m_2 \Delta y - 1.55 m_2 \Delta x)/r^2$,
which is a linear combination of the components of the true
force, $m_2 \hat{r}/r$. We can see this is approximately the
same linear transformation as the components in the left plot of
\cref{fig:square_truth2}, but this algebraic expression was learned
from scratch. 

We now test whether the GN will generalize to more nodes better than a GN with a larger $L^{e'}$.
This is because it is possible for a GN to ``cheat''
with a high dimension message-passing space, trained
on a fixed number of bodies. One example
of cheating would be for $\phi^e$ to
concatenate each sending node's properties along 
the message, and $\phi^v$ to calculate forces from these
and add them. When a new body is added, this calculation
might break.
While it is still possible for $\phi^e$
to develop an elaborate encoding scheme with $L^{e'}=D$
to cheat at this problem, it seems more natural for $\phi^e$
to learn the true force when $L^{e'}=D$ and therefore show improved generalization
to a greater number of nodes.

\begin{figure}[h]
    \centering
    \begin{subfigure}[b]{0.4\textwidth}
    \centering
    \includegraphics[width=\textwidth]{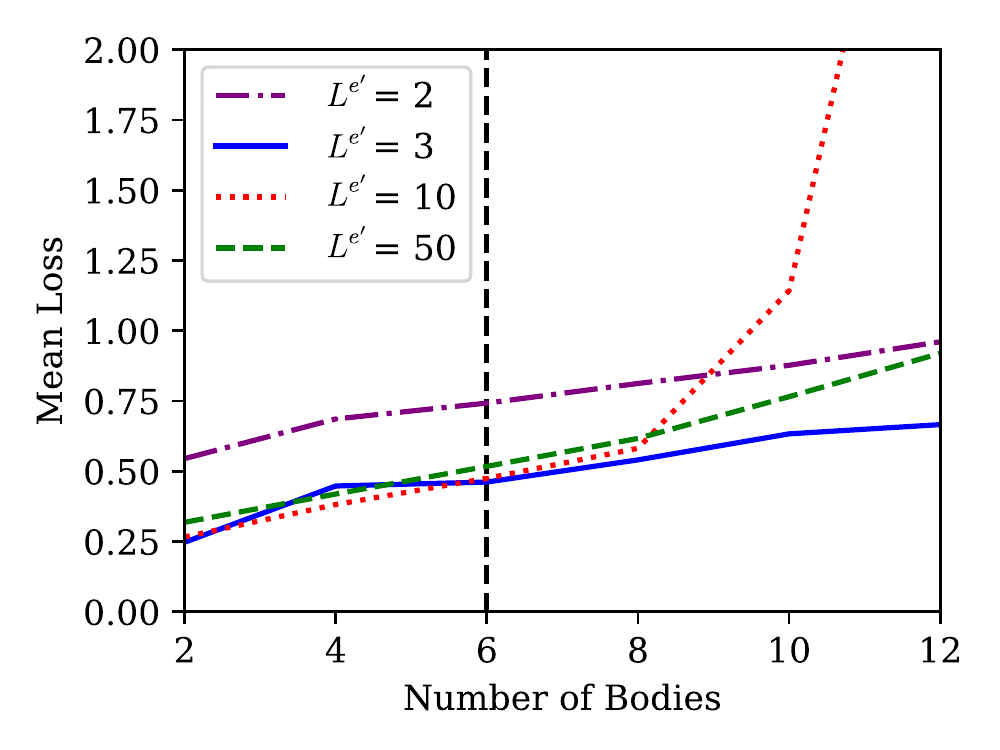}
    \end{subfigure}%
    \begin{subfigure}[b]{0.4\textwidth}
    \centering
    \includegraphics[width=\textwidth]{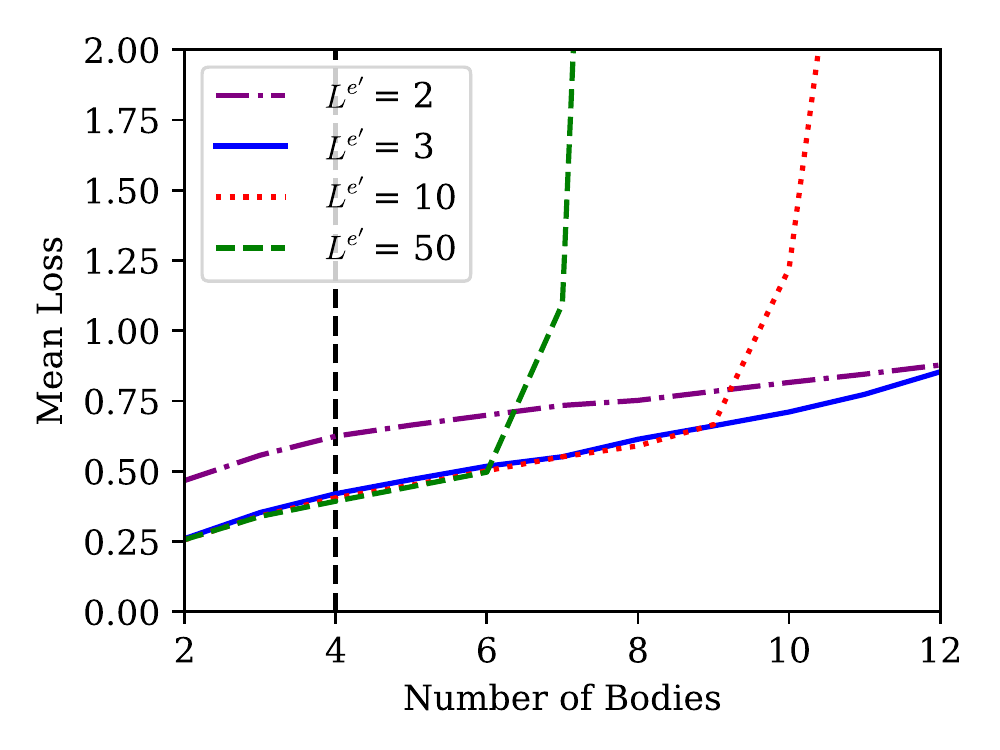}
    \end{subfigure}
    \caption{These plots demonstrate the improvement in generalization
    from minimizing the message passing space.
    The loss of GNs with different message-passing
    space dimension ($L^{e'}$), trained on a 6-body and 4-body system,
    in the left and right plots, respectively
    (indicated by the vertical line), are tested on a variable
    number of bodies in a $1/r^2$ simulation in 3D.}
   \label{fig:generalize}
\end{figure}

We test the hypothesis of better generalization
with $L^{e'}=D$ in \cref{fig:generalize},
by training GNs with different $L^{e'}$ on the 
3D $1/r^2$ simulations.
The observed trend is that systems with $L^{e'}>D$
see their loss blow up with a larger number of
bodies --- presumably because they have ``cheated'' slightly
and not learned the force law in $\phi^e$ but in a combination
of $\phi^e$ and $\phi^v$, whereas the $L^{e'}\in\{2, 3\}$ systems' $\phi^e$
has learned a projection of the true forces and is able to generalize better
for greater number of bodies.
A conclusion of this may be that one can optimize GNs
by minimizing $L^{e'}$ to the known minimum dimension
required to transmit information (e.g., $3$ for 3D forces), or, if this
dimension is unknown, until the loss drops off.

\section{Conclusion}

We have demonstrated an approach for imposing physically motivated inductive biases on graph networks to learn interpretable representations and improved zero-shot generalization. We have shown through experiment that our graph network models which implement this inductive bias can learn message representations equivalent to the true force vector for n-body gravitational and spring-like simulations in 2D and 3D. We also have demonstrated a generic technique for finding an unknown force law: symbolic regression models to fit explicit algebraic equations to our trained model's message function. Because GNs have more explicit sub-structure than their more homogeneous deep learning relatives (e.g., plain MLPs, convolutional networks), we can draw more fine-grained interpretations of their learned representations and computations. Finally, we have demonstrated that our model generalizes better at inference time to systems with more bodies than had been experienced during training.

{\it Acknowledgments:}\ Miles Cranmer and Rui Xu thank Professor S.Y. Kung for insightful suggestions on early work, as well as Zejiang Hou for his comments on an early presentation. Miles Cranmer would like to thank David Spergel for advice on this project, and Thomas Kipf, Alvaro Sanchez, and members of the DeepMind team for helpful comments on a draft of this paper. We thank the referees for insightful comments that both improved this paper and inspired future work.
\bibliographystyle{abbrvnat}
\bibliography{main}

\end{document}